\pdfoutput=1

\documentclass[11pt]{article}

\usepackage[]{ACL2023}

\usepackage{times}
\usepackage{latexsym}

\usepackage[T1]{fontenc}

\usepackage[utf8]{inputenc}
\usepackage{csquotes}

\usepackage{microtype}

\usepackage{inconsolata}
\usepackage{tabularx}
\usepackage{float}
\usepackage{caption}
\usepackage{graphicx}
\usepackage{multirow}
\usepackage{booktabs}
\usepackage{setspace}
\usepackage{booktabs}
\usepackage{hyperref}
\usepackage{paralist}
\usepackage{makecell}

\usepackage[normalem]{ulem}
\useunder{\uline}{\ul}{}

\title{Echoes of Discord: Forecasting Hater Reactions to Counterspeech}

\author{First Author \\
  Affiliation / Address line 1 \\
  Affiliation / Address line 2 \\
  Affiliation / Address line 3 \\
  \texttt{email@domain} \\\And
  Second Author \\
  Affiliation / Address line 1 \\
  Affiliation / Address line 2 \\
  Affiliation / Address line 3 \\
  \texttt{email@domain} \\}

\author{
  \textbf{Xiaoying Song\textsuperscript{1}}
  \textbf{Sharon Lisseth Perez\textsuperscript{1}}
  \textbf{Xinchen Yu\textsuperscript{2}}
  \textbf{Eduardo Blanco\textsuperscript{2}}
  \textbf{Lingzi Hong\textsuperscript{1}}
\\
  \textsuperscript{1}College of Information, University of North Texas
\\
  \textsuperscript{2}Department of Computer Science, University of Arizona
\\
  \small{
  \{xiaoyingsong, sharonperez\}@my.unt.edu} \\ 
  \small{
  \{xinchenyu, eduardoblanco\}@arizona.edu lingzi.hong@unt.edu  
  }
}

\begin{document}

\maketitle

\begin{abstract}
Hate speech (HS) erodes the inclusiveness of online users and propagates negativity and division. 
Counterspeech has been recognized as a way to mitigate HS. 
While some research has investigated the impact of user-generated counterspeech on social media platforms, few have examined and modeled haters' reactions toward counterspeech, despite the immediate alteration of haters' attitudes being an important aspect of counterspeech. 
This study fills the gap by analyzing the impact of counterspeech from the hater's perspective, focusing on whether the counterspeech leads the hater to reenter the conversation and if the reentry is hateful. 
We compile the \textbf{Re}ddit \textbf{Ec}h\textbf{o}es of Hate dataset (\textbf{ReEco}), which consists of triple-turn conversations featuring haters' reactions, to assess the impact of counterspeech. To predict haters' behaviors, we employ two strategies: a two-stage reaction predictor and a three-way classifier.
The linguistic analysis sheds insights on the language of counterspeech to hate eliciting different haters' reactions. 
Experimental results demonstrate that the 3-way classification model outperforms the two-stage reaction predictor, which first predicts reentry and then determines the reentry type.
We conclude the study with an assessment showing the most common error causes in the best-performing model. The dataset used in this study is publicly available for further research \footnote{\url{https://github.com/oliveeeee25/counterspeech_effectiveness_hater_reentry}}.

\end{abstract}

\textcolor{red}{Trigger warning: Read with caution. Examples in this paper may present toxic languages}

\section{Introduction}
Hate speech (HS) online causes increased prejudice and discrimination, fostering an environment of hostility and social division~\cite{chetty2018hate}. 
This motivates researchers to explore methods to mitigate the negative impact of HS, including the automatic detection of HS~\cite{abro2020automatic,duwairi2021deep} and generation of counterspeech~\cite{ saha2022countergedi, tekiroglu2022using}. 

Directly addressing HS through counterspeech is regarded as an effective strategy to curb the spread of hatred and promote constructive dialogue~\cite {bonaldi2022human}. 
Previous research has employed crowdsourced workers~\cite{qian2019benchmark} or NGO experts~\cite{chung2021towards} to craft counterspeech datasets, which are used to develop models for automatic counterspeech generation. 
However, the actual effect of synthetic counterspeech on social media platforms is unknown.

\begin{table}[t]
\centering
\large
\resizebox{0.48\textwidth}{!}{
\begin{tabular}{ m{0.3cm} m{12cm} }
\toprule
A & So you’re so big of a pu*** that you know something is wrong but the smell of a fast food restaurant makes it impossible for you to stop? Jesus Christ you’re a little b***.\\ \midrule
B & People like you are why vegans get bad rep, it doesn’t help to chastise the very people you’re trying to convert. I’d stay away from the evangelizing aspect of veganism and let others do that work because you’re clearly really not good at it, in fact people like you are detrimental to the cause. \\ \midrule
A & I don’t care. Stop being a b*** and doing something you know is wrong. I don’t have to coddle an abuser \\ 
\bottomrule
\end{tabular}}
\caption{An example of hater reentry from a Reddit conversation. User A posted a hate comment. User B replied to the hate speech, which induced A to reenter the conversation with a hateful post.}
\label{t:example}
\end{table}

Indeed, counterspeech may generate unfavorable conversation outcomes. Table~\ref{t:example} shows a slightly altered dialogue extracted from Reddit. 
The initial post by User A attacked someone with hateful words. 
In response, User B attempted to alter User A's perspective by stating, ``It doesn't help to...'' However, this reply angered User A, resulting in User A's reentry into the conversation with more hateful content.
While the intent was to counter hatred, the outcome was counterproductive—rather than de-escalating the situation, the response fueled further hostility, intensifying the conflict. 

\begin{figure}
\includegraphics[width=0.48\textwidth]{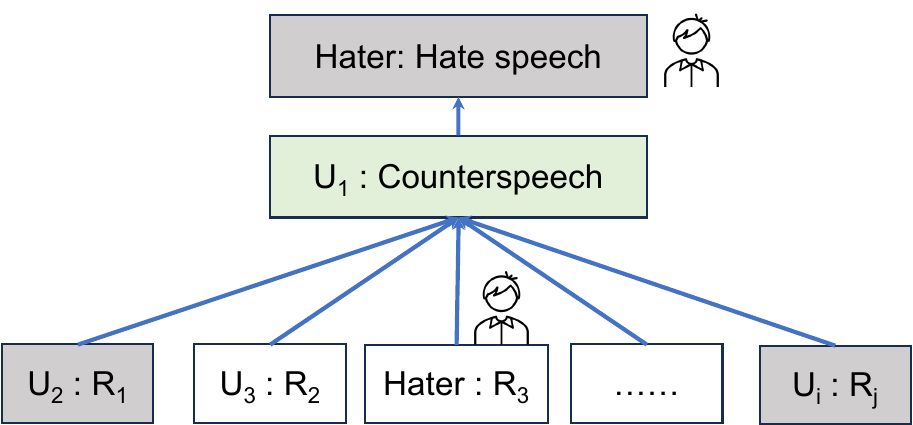} 
  \centering 
  \caption{Hater's non-hateful reentry as a conversation outcome. A Reddit user (hater) posts HS. Another user, $U_{1}$, replies with countersppech. This is followed by subsequent replies ($R_{1}$, $R_{2}$, ...$R_{j}$). The counterspeech prompts the hater to reenter the conversation with a non-hateful post. Grey boxes represent HS.}
  \label{Figure 1}
  \vspace{-.6cm}
\end{figure}
Several studies have proposed to assess the effectiveness of counterspeech by measuring conversation incivility~\cite{he2021racism,baider2023accountability,yu2023hate}. This measurement considers the reactions of all participants including bystanders.

Changing the attitudes and behaviors of haters is an important aspect of counterspeech~\cite{stroebe2008strategies}. However, few studies have investigated the impact of counterspeech from the perspective of hate perpetrators. 
Effective counterspeech can induce cognitive dissonance in haters by challenging their beliefs or assumptions, leading them to reconsider and potentially change behavior~\cite{dillard2002cognitive}. Counterspeech may make perpetrators aware of moral implications and the harmful impact of their actions, potentially re-reengaging their sense of empathy and leading to behavior change~\cite{wachs2022online}. On the other hand, hate perpetrators may perceive counterspeech as coercive, which triggers their stronger attachment to original beliefs and more resistance to change~\cite{acheme2024effects}. It is important to identify effective counterspeech to hate perpetrators: the ones that lead to immediate moderation of the hater's rhetoric~\cite{baider2023accountability,leonhard2018perceiving}, or at least would not incite the hater to be more hateful~\cite{shugars2019keep}. 

In this study, we investigate haters' reactions following counterspeech to HS in real conversations on a social media platform, Reddit. Our research addresses the following questions:
\begin{compactitem}
    \item Are there linguistic differences in counterspeech that lead to different hater reactions? 
    \item Which language models can more accurately predict hater reactions?
\end{compactitem}

Examining haters' reactions can reveal which characteristics of counterspeech are more likely to encourage constructive behavior. These insights can guide users to address HS in a way that minimizes the risk of provoking further negativity.
By developing models that predict haters' reactions, we can assess counterspeech, whether manually crafted by crowdsourced workers or generated by algorithms, and identify responses that are most likely to elicit desired hater reactions. 
The curated genuine conversation data can improve the training of large language models (LLMs) to generate counterspeech that effectively manages hater reactions.

To answer these questions, we build the \textbf{Re}ddit \textbf{Ec}h\textbf{o}es of Hate corpus (ReEco), which includes conversation pairs (HS and counterspeech) labeled by hater reactions: no reentry, hateful reentry, and non-hateful reentry. 
We perform linguistic analysis to identify variations in counterspeech that elicit different hater reactions.
We experiment with language models and adopt two strategies for predicting haters' reactions: \textbf{(i) a two-stage reaction predictor} that is composed of two predictive models: identifying hate reentry (yes/no) followed by identifying reentry type (hateful or non-hateful).
This method divides the prediction task into two stages, allowing each model to focus on a simpler, more defined problem. Considering that errors in the first model can affect the second model's input and lead to compounded inaccuracies \cite{lambert2022investigating}, we develop \textbf{(ii) a 3-way classifier} that predicts one of the three outcomes (no reentry, hateful reentry, or non-hateful reentry). 
Experiments show the 3-way classification achieves the highest prediction accuracy, while large language models (LLMs) are not superior in predicting haters' reactions compared to BERT models.

\section{Related Work}


\begin{table*}[ht]
\centering
\renewcommand{\arraystretch}{1.2} 
\scalebox{0.7}{
\begin{tabular}{lcccccc}
\hline
\multirow{2}{*}{\textbf{Community}} & 
\multirow{2}{*}{\# \textbf{Hate Speech}} & 
\multirow{2}{*}{\# \textbf{Counterspeech with Follow-up}} & 
\multicolumn{3}{c}{\textbf{Outcomes}} \\ 
\cline{4-6}
& & & \textbf{No Reentry (\%)} & \textbf{Hateful Reentry (\%)} & \textbf{Non-hateful Reentry (\%)} \\ 
\hline
\textbf{Discussion}      & 6,862  & 1,024  & 402 (39\%)  & 148 (15\%)  & 474 (46\%)  \\
\textbf{Identity}        & 7,779  & 2,247  & 721 (32\%)  & 468 (21\%)  & 1,058 (47\%)  \\
\textbf{Media-sharing}   & 2,549  & 1,417  & 435 (31\%)  & 309 (22\%)  & 673 (47\%)  \\
\textbf{Meme}           & 2,399  & 528    & 143 (27\%)  & 145 (27\%)  & 240 (46\%)  \\
\textbf{Hobby}          & 2,441  & 507    & 179 (35\%)  & 98 (19\%)   & 230 (46\%)  \\ 
\hline
\textbf{All}            & 22,030 & 5,723  & 1,880 (33\%) & 1,168 (20\%) & 2,675 (47\%) \\ 
\hline
\end{tabular}}
\caption{Analysis of our corpus, ReEco.}
\label{table2}
\end{table*}

\textbf{Counterspeech to HS} Counterspeech refers to a direct response to address HS~\cite{mathew2019thou}. Unlike approaches that block HS or haters, counterspeech fosters dialogues and promotes understanding without suppressing freedom of expression~\cite{schieb2016governing}, which can effectively de-escalate tension and reduce hostility~\cite{hangartner2021empathy}.
Previous studies have curated counterspeech datasets by employing humans or NGO experts to generate synthetic counterspeech~\cite{qian2019benchmark,chung2019conan}. Manual crafting of counterspeech is time-consuming and limits the capability of scaling.
Models have been developed for generating counterspeech~\cite{bonaldi2022human,chung2020italian,chung2021towards,zhu2021generate}. Few of these studies have considered the effectiveness of counterspeech from the perspective of conversation outcomes.
We bridge this gap by examining user-generated counterspeech and its impact on haters' reactions.

\noindent\textbf{Effectiveness of Counterspeech} 
Assessing the effectiveness of counterspeech is crucial for crafting appropriate counterspeech that mitigates the harms of HS~\cite{cepollaro2023counterspeech}. 
\citet{baider2023accountability} uses the number of answers triggered by the comment and the tone of the answers to determine the effectiveness of counterspeech. 
\citet{yu2023hate} assess the effectiveness based on conversation incivility, which considers the number of hateful and non-hateful comments, along with the unique authors participating.
Most of these works evaluate the effectiveness based on bystanders of the conversation.
\citet{reynolds2016counter} propose using qualitative analysis to assess the effect of counterspeech based on the hate perpetrators. 
This study is the first to model haters' reactions with genuine conversation data from a social media platform. 

\noindent\textbf{Conversation Outcome Forecasting} Methods have been developed to predict the future development of a conversation~\cite{bao2021conversations}. Previous works predict conversational outcomes from different aspects, such as users’ reentry~\cite{wang2021re,zeng2019joint}, the popularity of a conversation~\cite{rizos2016predicting,risch2020top,backstrom2013characterizing}, conversation killers \cite{jiao2018find}, and derailment of conversations \cite{chang2019trouble}. In particular, \citet{backstrom2013characterizing} has used users' network relationships, and \citet{zeng2019joint} has used users' historical behaviors to predict reentry behaviors. However, these studies do not provide direct insights into how counterspeech influences subsequent conversation outcomes, which are crucial for developing counterspeech generation strategies and models. Additionally, few studies have modeled haters' reactions following counterspeech.

\section{A New HS/counterspeech Corpus - ReEco}
\textbf{Data Collection and Process} 
We use the PushShift API to collect complete conversation threads containing HS from Reddit.\footnote{\url{https://pushshift.io/api-parameters/}} We employ community-based sampling, selecting 42 subreddits that are identified to exhibit a higher prevalence of HS by \citet{qian2019benchmark}, \citet{guest2021expert}, and \citet{vidgen2019challenges}, including r/MensRights, r/PurplePillDebate, and r/Seduction, etc (see Appendix \ref{Subreddit List} for the complete list). We collect a total of 1,410,361 comments ranging from February 2, 2019, to November 26, 2021.

\noindent\textbf{Detect HS} HS refers to expressions in which the author deliberately targets an individual or group with the intent to vilify, humiliate, or incite hatred \cite{yu2022hate}. 
We fine-tune RoBERTa models~\cite{liu2019roberta} with the HS datasets by ~\citet{vidgen2021introducing},~\citet{qian2019benchmark}, and \citet{davidson2017automated}, and build three HS classifiers. A comment is labeled as hateful only if all three classifiers consistently predict it as such. This is to minimize the risk of any classifiers being wrong.
The prediction-assigned labels are compared with human annotations to validate reliability. Two researchers manually annotated 200 samples guided by the HS definition and examples (annotation details in Appendix \ref{Annotation Details}). The agreement rate is 93\%. The Kappa coefficient between the predicted results and manual labeling is $0.76$, indicating a substantial level of agreement~\cite{viera2005understanding}.
The manual evaluation verifies the accuracy of the predictions. While the classifiers are not flawless, most of the conversations identified are related to HS, ensuring the generation of reliable results.


\begin{table*}[t]
    \centering
    \begin{minipage}{0.49\textwidth}
        \centering
        \renewcommand{\arraystretch}{1}
        \resizebox{\textwidth}{!}{
        \setlength{\tabcolsep}{.075in}
        \large
        \begin{tabular}{@{}lcccccl@{}}
        \toprule
                                      & Meme                   & Hobby                  & Identity               & Discussion             & Media-sharing          & All          \\ \midrule
        \textbf{Textual Factors}      &                        &                        &                        &                        &                        &              \\
        ~~~~2nd Person Pronouns       & $\uparrow$             & $\uparrow$             & $\uparrow$             & $\uparrow$             & $\uparrow$             & $\uparrow$   \\
        ~~~~Uncertainty               & $\underline\uparrow$   & $\underline\uparrow$   & $\underline\uparrow$   & $\uparrow$             & $\underline\uparrow$   & $\uparrow$   \\
         ~~~~Abstract                  & $\underline\uparrow$   & $\underline\downarrow$ & $\underline\uparrow$   & $\uparrow$             & $\underline\uparrow$   & $\uparrow$   \\ \midrule
        \textbf{Emotional Factors}    &                        &                        &                        &                        &                        &              \\
        ~~~~Enlightenment             & $\uparrow$             & $\underline\uparrow$   & $\uparrow$             & $\uparrow$             & $\underline\uparrow$   & $\uparrow$   \\
        ~~~~Negative                  & $\underline\uparrow$   & $\underline\downarrow$ & $\underline\uparrow$   & $\underline\uparrow$   & $\underline\uparrow$   & $\uparrow$   \\
        ~~~~Fear                      & $\underline\uparrow$   & $\underline\downarrow$ & $\uparrow$             & $\underline\uparrow$   & $\underline\uparrow$   & $\uparrow$   \\
        ~~~~Positive                  & $\underline\downarrow$ & $\underline\downarrow$ & $\underline\downarrow$ & $\underline\downarrow$ & $\underline\downarrow$ & $\downarrow$ \\ 
        ~~~~Polarity                  & $\underline\uparrow$   & $\underline\downarrow$ & $\uparrow$             & $\uparrow$             & $\uparrow$             & $\uparrow$   \\
        ~~~~Valence                   & $\underline\uparrow$   & $\underline\uparrow$   & $\underline\uparrow$   & $\uparrow$             & $\underline\uparrow$   & $\uparrow$   \\ \bottomrule
        \end{tabular}}
        \caption{Linguistic analysis comparing the counterspeech that triggers hater reentry and no reentry in different communities. The up arrow indicates higher values in counterspeech with hater reentry. Tests that do not pass the Bonferroni correction are underlined.}
        \label{lingusitic analysis: reentry}
    \end{minipage}
    \hfill 
    \begin{minipage}{0.49\textwidth}
        \centering
        \resizebox{\textwidth}{!}{
        \begin{tabular}{@{}lcccccc@{}}
        \toprule
                                      & Meme                   & Hobby                  & Identity               & Discussion             & Media-sharing          & All          \\ \midrule
        \textbf{Textual Factors}      &                        &                        &                        &                        &                        &              \\
        ~~~~Causation                 & $\underline\uparrow$   & $\underline\uparrow$   & $\uparrow$             & $\underline\uparrow$   & $\underline\uparrow$   & $\uparrow$   \\
        ~~~~Format                    & $\underline\uparrow$   & $\underline\uparrow$   & $\uparrow$             & $\underline\uparrow$   & $\underline\uparrow$   & $\uparrow$   \\ \midrule
        \textbf{Social-related Factors}&                        &                        &                        &                        &                        &              \\
        ~~~~Respect                   & $\underline\downarrow$ & $\underline\downarrow$ & $\downarrow$           & $\underline\downarrow$ & $\underline\downarrow$ & $\downarrow$ \\
        ~~~~Power                     & $\underline\downarrow$ & $\underline\downarrow$ & $\underline\downarrow$ & $\downarrow$           & $\underline\downarrow$ & $\downarrow$ \\
        ~~~~Worship                   & $\underline\downarrow$ & $\underline\downarrow$ & $\underline\downarrow$ & $\underline\downarrow$ & $\underline\downarrow$ & $\downarrow$ \\
        ~~~~Forgiveness               & $\underline\downarrow$ & $\underline\downarrow$ & $\underline\downarrow$ & $\underline\downarrow$ & $\underline\downarrow$ & $\underline\downarrow$  \\ \midrule
        \textbf{Emotional Factors}     &                        &                        &                        &                        &                        &              \\
        ~~~~Longing                   & $\underline\downarrow$ & $\underline\downarrow$ & $\underline\downarrow$ & $\downarrow$           & $\downarrow$           & $\downarrow$ \\
        ~~~~Exclamation               & $\underline\uparrow$   & $\underline\uparrow$   & $\uparrow$             & $\underline\uparrow$   & $\underline\uparrow$   & $\uparrow$   \\
        ~~~~Aggression                & $\underline\uparrow$   & $\underline\uparrow$   & $\uparrow$             & $\underline\uparrow$   & $\underline\uparrow$   & $\uparrow$   \\ \bottomrule
        \end{tabular}}
        \caption{Linguistic analysis comparing the counterspeech that triggers hateful/non-hateful reentry in different communities. The up arrow indicates higher values in counterspeech with hateful reentry. Tests that do not pass the Bonferroni correction are underlined.}
        \label{lingusitic analysis: reentry type}
    \end{minipage}
\end{table*}

\noindent\textbf{Identify Counterspeech} Counterspeech are responses to HS which are crafted to contradict and challenge hateful remarks \cite{chung2023understanding}.
For comments predicted to be HS, we obtain their subsequent two-level replies. There are 40,162 HS and 76,009 replies to these HS. Of these pairs, only 22,030 HS and their replies have follow-up replies, which are used to analyze hater reactions. 
We exclude pairs without follow-up replies as these pairs are located at the end of the dialogue tree, which could happen when there is topic exhaustion after thorough discussions or external interruptions~\cite{jiao2018find}. By excluding these pairs, we can largely avoid the impact of these factors and focus on haters' `no reentry' when the conversations are actively evolving.  

Not all replies to HS qualify as counterspeech. We developed three classifiers, fine-tuned on three distinct counterspeech datasets~\cite{albanyan2023finding, yu2022hate, vidgen2021introducing}, to identify counterspeech within those replies.
We consider a reply to be counterspeech if the three classifiers indicate so, which results in 5,723 (HS, counterspeech) pairs from Reddit. We manually annotated 200 samples to validate the results (annotation details in Appendix \ref{Annotation Details}). The agreement rate is 95\%. The Kappa coefficient between the predicted results and manual labeling is $0.79$, indicating the result is reliable~\cite{viera2005understanding}.
We examine whether the user who posts the initial HS (i.e., hater) shows up in the follow-up conversation to counterspeech, categorizing the conversation as with hater `reentry' or `no reentry.' Based on the prediction of HS classifiers, the reentry comments are further labeled as `hateful' or `non-hateful'.
Figure~\ref{Figure 1} illustrates the outcome of a conversation that involves a hater's reentry with a non-hateful reply to counterspeech.

\noindent\textbf{Description of ReEco}
The final corpus, ReEco, consists of 5,723 (HS, counterspeech) pairs. To investigate whether the hater reentry behavior varies across communities, we group the 42 subreddits into 5 categories (see Appendix \ref{Subreddit List}), i.e., \textit{Discussion}, \textit{Identity}, \textit{Media-sharing}, \textit{Meme}, and \textit{Hobby}~\cite{weld2022makes}, and analyze the distribution of hater reentry labels in Table~\ref{table2}. 
There are more HS in \textit{Discussion} and \textit{Identity}. 
Different communities exhibit similar proportions of \textit{non-hateful reentry} ($\approx 46\%$), \textit{hateful reentry} ($\approx 21\%$), and \textit{no reentry} ($\approx 33\%$). However, there are notable variations. The \textit{Meme} community shows a slightly higher proportion of \textit{hateful reentries} (27\%) and a lower proportion of \textit{no reentry} (27\%), whereas the \textit{Discussion} community has a higher proportion of \textit{no reentries} (39\%) and a lower proportion of \textit{hateful reentries} (15\%).

\begin{table*}[ht!]
    \centering
    \scalebox{0.75}{
    \begin{tabular}{l ccc ccc ccc}
        \toprule
        & \multicolumn{3}{c}{\textbf{Reentry}} & \multicolumn{3}{c}{\textbf{No Reentry}}  & \multicolumn{3}{c}{\textbf{Weighted Average}} \\
        \cmidrule(lr){2-4} \cmidrule(lr){5-7} \cmidrule(lr){8-10}  
        \textbf{Model} & \textbf{P} & \textbf{R} & \textbf{F1} & \textbf{P} & \textbf{R} & \textbf{F1} & \textbf{P} & \textbf{R} & \textbf{F1}  \\
        \midrule
        \addlinespace
        \textbf{Baseline}  & 0.69 & 1.00 & 0.82 & 0.00 & 0.00 & 0.00 & 0.48 & 0.69 & 0.57 \\
        \addlinespace
        \textbf{BERT} & & & & & & & & & \\
        \hspace{3mm} HS & 0.77 & 0.69 & 0.73 & 0.72 & 0.80 & 0.75 & 0.74 & 0.74 & 0.74 \\
        \hspace{3mm} Counterspeech & 0.87 & 0.71 & 0.78 & 0.75 & 0.89 & 0.82 & 0.81 & 0.80 & 0.80 \\
        \hspace{3mm} Pair & 0.89 & 0.69 & 0.78 & 0.75 & 0.91 & 0.82 & 0.82 & 0.80 & 0.80 \\
        \addlinespace
        \textbf{BERT-MTL} & & & & & & & & & \\
        \hspace{3mm} HS & 0.79 & 0.75 & 0.77 & 0.76 & 0.80 & 0.78 & 0.78 & 0.78 & 0.78 \\
        \hspace{3mm} Counterspeech & 0.84 & 0.78 & 0.81 & 0.79 & 0.85 & 0.82 & 0.81 & 0.81 & 0.81 \\
        \hspace{3mm} Pair & 0.86 & 0.79 & \textbf{0.82} & 0.81 & 0.87 & \textbf{0.84} & 0.83 & 0.83 & \textbf{0.83} \\
        \addlinespace
        \textbf{LLaMA 3 Zero-shot} & & & & & & & & & \\
        \hspace{3mm} HS & 0.39 & 0.83 & 0.53 & 0.55 & 0.14 & 0.22 & 0.49 & 0.42 & 0.35 \\
        \hspace{3mm} Counterspeech & 0.60 & 0.51 & 0.55 & 0.40 & 0.49 & 0.44 & 0.50 & 0.50 & 0.49 \\
        \hspace{3mm} Pair & 0.41 & 0.60 & 0.49 & 0.62 & 0.43 & 0.50 & 0.53 & 0.50 & 0.50 \\
        \addlinespace
        \textbf{LLaMA 3 Finetuned} & & & & & & & & & \\
        \hspace{3mm} HS & 0.74 & 0.83 & 0.78 & 0.41 & 0.29 & 0.33 & 0.64 & 0.67 & 0.65 \\
        \hspace{3mm} Counterspeech & 0.73 & 0.83 & 0.78 & 0.40 & 0.27 & 0.33 & 0.64 & 0.67 & 0.65 \\
        \hspace{3mm} Pair & 0.75 & 0.85 & 0.80 & 0.46 & 0.30 & 0.36 & 0.66 & 0.69 & 0.67 \\
        \bottomrule
    \end{tabular}
    }
    \caption{Performance comparison of different models and inputs on reentry prediction.}
    \label{tab:hate_speech_reentry}
    \vspace{-.4cm}
\end{table*}

\section{Corpus Analysis}
We conduct two linguistic analyses, the first one analyzes differences in counterspeech that elicit hater to reenter or not. 
The second one takes the counterspeech with hater reentry and investigates the differences that elicit hateful or non-hateful reentry. 
These analyses are carried out across different communities, as topics and sentiments may vary in communities. 
We use SEANCE ~\cite{crossley2017sentiment} for sentiment analysis and spaCy ~\cite{honnibal2017spacy} for entity recognition. 
We further employ the Wilcoxon rank-sum test to perform statistical comparisons. The Bonferroni correction is applied to highlight the most significant linguistic features ~\cite{weisstein2004bonferroni}. (See details of linguistic features in Appendix \ref{Linguistic Factors})

Table~\ref{lingusitic analysis: reentry} details linguistic findings in counterspeech that trigger haters' reentry. 
Counterspeech that includes elements of enlightenment, and negative and fear emotions is more likely to trigger hater reentry, which is consistent across most communities, except Meme. Counterspeech with hater reentry (except Hobby) has significantly more polarity and valence words, indicators of emotionally charged and positioned language. This phenomenon can be explained by the emotional contagion theory~\cite{hancock2008m} and the affective intelligence theory~\cite{marcus2011parsimony}. Emotions in counterspeech can lead to haters feeling and displaying similar emotions. These emotions influence how the hater processes information and engages in the conversation~\cite{choi2021enthusiasm}.

We observe some new findings that have been seldom mentioned or differ. 
Second-person pronounces, uncertainty (words denoting feelings of uncertainty, e.g., about, almost), and abstract words (e.g., words reflecting a tendency to use abstract vocabulary, e.g., ability, advantage) show higher prevalence in counterspeech that causes hater reentry except in Hobby. 
Unlike the findings of \cite{lubis2019positive}, positive emotions in counterspeech tend to inhibit further engagement from haters (i.e., significantly more in no reentry). This suggests counterspeech has distinctive characteristics compared to general conversational engagement.

These observations imply that the emotional tone of counterspeech can influence haters' reactions. \citet{ghandeharioun2019towards} have proposed designing emotionally aware agents to promote behavior change in wellness applications. Our analysis suggests this idea may also have value in altering haters' behaviors. 

Table~\ref{lingusitic analysis: reentry type} presents the linguistic differences between counterspeech that causes different types of reentry. 
Significant differences are summarized into three aspects, including textual, social-related, and emotional factors. 
Textual factors, such as causation and format, occur more frequently in instances of hateful reentry and this trend is consistent across all communities. 
In social-related factors, counterspeech that prompts non-hateful reentry tends to include more elements of respect, power, worship, or forgiveness. 
These insights provide clues to encourage positive interactions and facilitate constructive dialogue. 
It is hypothesized that haters modify their behavior in such cases due to cognitive dissonance, as continuing to engage in hateful actions while receiving respect~\cite{dillard2002cognitive} and mutual understanding~\cite{clark2019makes} may create psychological discomfort. Further research is needed to explore the underlying mechanisms behind how these signals prompt behavioral change in haters.

\begin{table*}[t]
    \centering
    \scalebox{0.75}{
    \begin{tabular}{l ccc ccc ccc}
        \toprule
        & \multicolumn{3}{c}{\textbf{Hateful Reentry}} & \multicolumn{3}{c}{\textbf{Non-Hateful Reentry}}  & \multicolumn{3}{c}{\textbf{Weighted Average}} \\
        \cmidrule(lr){2-4} \cmidrule(lr){5-7} \cmidrule(lr){8-10}  
        \textbf{Model} & \textbf{P} & \textbf{R} & \textbf{F1} & \textbf{P} & \textbf{R} & \textbf{F1} & \textbf{P} & \textbf{R} & \textbf{F1}  \\
        \midrule
        \addlinespace
        \textbf{Baseline}  & 0.00 & 0.00 & 0.00 & 0.70 & 1.00 & 0.82 & 0.48 & 0.70 & 0.57 \\
        \addlinespace
        \textbf{BERT} & & & & & & & & & \\
        \hspace{3mm} HS & 0.67 & 0.83 & 0.74 & 0.77 & 0.59 & 0.67 & 0.72 & 0.71 & 0.71 \\
        \hspace{3mm} Counterspeech & 0.68 & 0.91 & 0.78 & 0.86 & 0.57 & 0.69 & 0.77 & 0.74 & 0.73 \\
        \hspace{3mm} Pair & 0.74 & 0.89 & 0.81 & 0.87 & 0.68 & 0.76 & 0.80 & 0.79 & 0.78 \\
        \addlinespace
        \textbf{BERT-MTL} & & & & & & & & & \\
        \hspace{3mm} HS & 0.77 & 0.86 & 0.81 & 0.84 & 0.75 & 0.79 & 0.81 & 0.80 & 0.80 \\
        \hspace{3mm} Counterspeech & 0.76 & 0.87 & 0.81 & 0.85 & 0.73 & 0.78 & 0.80 & 0.80 & 0.80 \\
        \hspace{3mm} Pair & 0.80 & 0.84 & \textbf{0.82} & 0.84 & 0.79 & \textbf{0.81} & 0.82 & 0.82 & \textbf{0.82} \\
        \addlinespace
        \textbf{LLaMA 3 Zero-shot} & & & & & & & & & \\
        \hspace{3mm} HS & 0.59 & 0.86 & 0.70 & 0.42 & 0.14 & 0.21 & 0.52 & 0.57 & 0.50 \\
        \hspace{3mm} Counterspeech & 0.60 & 0.65 & 0.62 & 0.44 & 0.38 & 0.41 & 0.53 & 0.54 & 0.53 \\
        \hspace{3mm} Pair & 0.60 & 0.88 & 0.71 & 0.45 & 0.14 & 0.22 & 0.54 & 0.58 & 0.51 \\
        \addlinespace
        \textbf{LLaMA 3 Finetuned} & & & & & & & & & \\
        \hspace{3mm} HS & 0.32 & 0.22 & 0.26 & 0.70 & 0.79 & 0.74 & 0.58 & 0.61 & 0.59 \\
        \hspace{3mm} Counterspeech & 0.36 & 0.27 & 0.31 & 0.71 & 0.80 & 0.75 & 0.61 & 0.63 & 0.62 \\
        \hspace{3mm} Pair & 0.40 & 0.16 & 0.23 & 0.71 & 0.90 & 0.79 & 0.61 & 0.67 & 0.62 \\
        \bottomrule
    \end{tabular}
    }
    \caption{Performance comparison of different models and inputs on reentry type prediction.}
    \label{tab:reentry_type_prediction}
    \vspace{-.4cm}
\end{table*}

Regarding emotional factors, counterspeech with hateful reentry conveys elements like exclamation and aggression, while counterspeech with non-hateful reentry contains more longing emotion. It implies counterspeech with aggression potentially incites further hostility while expressions of longing for better understanding foster a less negative re-engagement. This further verifies that emotion in counterspeech plays a crucial role in shaping haters' reactions ~\cite{hancock2008m}. 





\section{Experiments and Results}
We experiment with two strategies to predict haters' reactions following counterspeech. The first \textbf{Two-stage Reaction Predictor} includes two consecutive tasks: (i) predicting whether a hater reenters the conversation, and (ii) predicting whether the reentry is hateful or not, if reentry occurs. The second \textbf{3-way Response Classifier} predicts the hater's reaction as one of the three results: no reentry, hateful reentry, and non-hateful reentry.

\subsection{Experiment Implementation}
We split ReEco into training (80\%) and testing (20\%), and report model performance on the test data. For each prediction task, we experiment with the following three types of input to identify which information can best model haters' reactions.

\noindent\textbf{HS} \citet{liu2018forecasting} found that hateful comments can predict future engagement in conversations. We adopt this idea and experiment with predicting haters' reactions based on the HS. 

\noindent\textbf{Counterspeech} How people respond to HS may elicit varying responses by the hater~\cite{shugars2019keep}. We explore how effectively counterspeech can predict haters' reactions.

\noindent\textbf{Pair} We hypothesize that the interaction between HS and counterspeech influences the hater's subsequent behavior. To model this interaction, we concatenate the HS and counterspeech using the [SEP] special token as input to BERT-large-uncased. For the LLM setup, the conversation pair is formed by concatenating the text in the query.

We experiment with the following models for prediction tasks. This is to validate the robustness of our findings, identify the most predictive information, and determine the optimal language models for accurate prediction.

\noindent\textbf{BERT} 
The model's neural architecture is built on a BERT-large-uncased base, followed by a fully connected layer with 1,024 neurons and tanh activation. This is then connected to a final fully connected layer with 3 neurons and softmax activation.

\noindent\textbf{Multi-task Learning (MTL) Models}
MTL allows a model to learn from multiple related tasks simultaneously. 
By integrating both shared and context-specific representations, the model could potentially achieve enhanced performance~ \cite{zhang2018overview}. 
We utilized a deep neural network architecture with a shared base, BERT-large-uncased, followed by task-specific layers for each classification output. 
The model is trained on three pertinent tasks: classification of toxicity,\footnote{Available publicly at https://www.kaggle.com/c/jigsaw-unintended-bias-in-toxicity-classification/data} detection of HS and counterspeech~\cite{he2021racism}, and identification of personal attacks~\cite{zhang2018conversations}.

\begin{table*}[ht!]
    \centering
    \scalebox{0.70}{
    \begin{tabular}{l ccc ccc ccc ccc}
        \toprule
        & \multicolumn{3}{c}{\textbf{Hateful Reentry}} & \multicolumn{3}{c}{\textbf{Non-Hateful Reentry}} & \multicolumn{3}{c}{\textbf{No Reentry}} & \multicolumn{3}{c}{\textbf{Weighted Average}} \\
        \cmidrule(lr){2-4} \cmidrule(lr){5-7} \cmidrule(lr){8-10} \cmidrule(lr){11-13}  
        \textbf{Model} & \textbf{P} & \textbf{R} & \textbf{F1} & \textbf{P} & \textbf{R} & \textbf{F1} & \textbf{P} & \textbf{R} & \textbf{F1} & \textbf{P} & \textbf{R} & \textbf{F1} \\
        \midrule
        \addlinespace
        \textbf{Baseline}  & 0.00 & 0.00 & 0.00 & 0.48 & 1.00 & 0.65 & 0.00 & 0.00 & 0.00 & 0.23 & 0.48 & 0.31 \\
        \addlinespace
        \textbf{BERT} & & & & & & & & & & & & \\
        \hspace{3mm} HS & 0.73 & 0.85 & 0.79 & 0.66 & 0.53 & 0.59 & 0.74 & 0.77 & 0.76 & 0.71 & 0.72 & 0.71 \\
        \hspace{3mm} Counterspeech & 0.81 & 0.80 & 0.80 & 0.66 & 0.52 & 0.58 & 0.67 & 0.73 & 0.71 & 0.71 & 0.71 & 0.71 \\
        \hspace{3mm} Pair & 0.73 & 0.89 & 0.80 & 0.70 & 0.55 & 0.61 & 0.80 & 0.80 & \textbf{0.80} & 0.74 & 0.75 & 0.74 \\
        \addlinespace
        \textbf{BERT-MTL} & & & & & & & & & & & & \\
        \hspace{3mm} HS & 0.81 & 0.83 & 0.82 & 0.68 & 0.56 & 0.62 & 0.71 & 0.82 & 0.76 & 0.74 & 0.71 & 0.73 \\
        \hspace{3mm} Counterspeech & 0.84 & 0.80 & 0.82 & 0.67 & 0.63 & 0.65 & 0.72 & 0.79 & 0.76 & 0.74 & 0.74 & 0.74 \\
        \hspace{3mm} Pair & 0.81 & 0.86 & \textbf{0.84} & 0.69 & 0.67 & \textbf{0.68} & 0.78 & 0.78 & 0.79 & 0.77 & 0.77 & \textbf{0.77} \\
        \addlinespace
        \textbf{LLaMA 3 Zero-shot} & & & & & & & & & & & & \\
        \hspace{3mm} HS & 0.44 & 0.10 & 0.16 & 0.28 & 0.44 & 0.35 & 0.39 & 0.57 & 0.46 & 0.38 & 0.35 & 0.31 \\
        \hspace{3mm} Counterspeech & 0.44 & 0.15 & 0.23 & 0.29 & 0.42 & 0.34 & 0.33 & 0.46 & 0.39 & 0.36 & 0.33 & 0.31 \\
        \hspace{3mm} Pair & 0.40 & 0.68 & 0.51 & 0.26 & 0.18 & 0.21 & 0.35 & 0.17 & 0.23 & 0.35 & 0.37 & 0.33 \\
        \addlinespace
        \textbf{LLaMA 3 Finetuned} & & & & & & & & & & & & \\
        \hspace{3mm} HS & 0.33 & 0.15 & 0.20 & 0.50 & 0.63 & 0.56 & 0.36 & 0.35 & 0.36 & 0.42 & 0.44 & 0.42 \\
        \hspace{3mm} Counterspeech & 0.35 & 0.17 & 0.23 & 0.52 & 0.66 & 0.58 & 0.42 & 0.37 & 0.39 & 0.45 & 0.47 & 0.45 \\
        \hspace{3mm} Pair & 0.37 & 0.17 & 0.24 & 0.52 & 0.67 & 0.59 & 0.42 & 0.39 & 0.40 & 0.46 & 0.48 & 0.46 \\
        \addlinespace
        \textbf{Two-stage (BERT-MTL)} & & & & & & & & & & & & \\
        \hspace{3mm} Pair & 0.13 & 0.27 & 0.17 & 0.73 & 0.75 & 0.74 & 1.00 & 0.26 & 0.42 & 0.70 & 0.51 & 0.53 \\
        \bottomrule
    \end{tabular}
    }
    \caption{Performance comparison of different models and inputs for 3-way reentry prediction.}
    \label{tab:three_way_prediction}
    \vspace{-.4cm}
\end{table*}

\noindent\textbf{Llama 3 Zero-shot} 
Large language models (LLMs), such as Llama 3, have been trained on large corpora and demonstrated capabilities in understanding human language.
We experiment with Llama 3 in dialogue to test whether the zero-shot learning model can accurately predict haters' reactions. To constrain the format of the generated response, we adopt Guidance, a programming paradigm that enables constrained generation with regex,\footnote{https://github.com/guidance-ai/guidance} on the Llama-3.1-8B-Instruct model for inferences.

The query for the reentry prediction is formulated as:
\begin{displayquote}
\footnotesize
\texttt{Here is a counterspeech to a hate comment: <example>. }\\
\texttt{Will the hater come back to join the conversation? Answer `Yes' or `No'. }
\end{displayquote}
The query for reentry type prediction is: 
\begin{displayquote}
\footnotesize
\texttt{Here is a counterspeech to a hate comment: <example>.}\\
\texttt{Assuming the hater comes back to join the conversation, will the engagement be hateful? Answer `Yes' or `No'}
\end{displayquote}
The query for the 3-way response is: 
\begin{displayquote}
\footnotesize
\texttt{Here is a counterspeech to a hate comment:<example>. }\\
\texttt{What will be the hater's response? Answer `Reentry with a non-hateful comment', `Reentry with a hateful comment', or `No reentry'}
\end{displayquote}
In addition to counterspeech, we also experiment with HS and the dialogue (HS+Counterspeech) to predict outcomes. For dialogue, the input for <example> is "hate comment:" + <hate comment> + "counterspeech:" + <counterspeech>.

\noindent\textbf{Llama 3 Finetuned} LLMs can be optimized for a specific task through fine-tuning. By feeding an LLM with training data, the model learns the linguistic patterns in counterspeech to HS with different hater reentry behaviors. We use texts and their labels in training data to construct dialogues as input to train the Llama-3.1-8B-Instruct model. Specifically, we apply the Low-Rank Adaptation (LoRA) ~\cite{hu2021lora} method in finetuning (details of model parameters in Appendix \ref{Hyperparameters of Models}). The trained models are then utilized to make predictions on the test data.

\subsection{Results}

\textbf{Hater Reentry Prediction}
This task predicts whether there is hater reentry. The majority baseline is calculated based on all samples predicted to be ``reentry''. Table~\ref{tab:hate_speech_reentry} shows the performance of models. We observe the following insights:

Most models perform significantly better than the baseline, except Llama 3 zero-shot learning. This suggests the reentry behavior of haters is predictable and LLM is limited in forecasting conversation outcomes. The performance of fine-tuned Llama 3 (best F1: 0.67) is also inferior to BERT (best F1: 0.80) and BERT-MTL (best F1: 0.83). 

HS/counterspeech pairs have the best predictive power across models, indicating the hater's reentry is not solely dependent on the HS or the counterspeech. 
Haters' reentry can be better predicted with counterspeech than HS by
BERT (counterspeech  (0.80) vs hate (0.74)), BERT-MTL models (counterspeech (0.81) vs hate (0.78)), and Llama 3 zero-shot (counterspeech (0.49) vs hate (0.35)).
It indicates that counterspeech is important in determining whether a hater will re-engage in the follow-up conversations.

\begin{table*}[t]
\centering
\small
\renewcommand{\arraystretch}{1.2}
\setlength{\tabcolsep}{8pt}
\begin{tabular}{lccccccc}
\toprule
\multirow{2}{*}{\textbf{Error Cause}} & \multicolumn{2}{c}{\textbf{Non-Hateful Reentry}} & \multicolumn{2}{c}{\textbf{Hateful Reentry}} & \multicolumn{2}{c}{\textbf{No Reentry}} & \multirow{2}{*}{\textbf{All}} \\ 
\cmidrule(lr){2-3} \cmidrule(lr){4-5} \cmidrule(lr){6-7}
& \textbf{FP} & \textbf{FN} & \textbf{FP} & \textbf{FN} & \textbf{FP} & \textbf{FN} &  \\ \midrule
\textbf{Rhetorical Questions} & 0.52 & 0.47 & 0.38 & 0.63 & 0.58 & 0.41 & 0.49 \\
\textbf{Negation} & 0.22 & 0.26 & 0.29 & 0.17 & 0.20 & 0.26 & 0.23 \\
\textbf{Sarcasm or Irony} & 0.00 & 0.21 & 0.27 & 0.09 & 0.16 & 0.20 & 0.17 \\
\textbf{Intricate Text} & 0.16 & 0.00 & 0.06 & 0.09 & 0.04 & 0.13 & 0.10 \\
\textbf{General Knowledge} & 0.10 & 0.06 & 0.00 & 0.02 & 0.02 & 0.00 & 0.01 \\ 
\bottomrule
\end{tabular}
\caption{Distribution of common errors in the best 3-way prediction model across different classes. The values represent the proportion of error causes. FP refers to false positives, and FN indicates false negatives.}
\label{error_analysis}
\vspace{-0.4cm}
\end{table*}

BERT-MTL achieves the best performance (F1: 0.83). The F1 score is significantly higher than the best BERT (F1: 0.80) and Llama 3 Finetuned (F1: 0.67) based on McNemar’s tests ($p<0.001$) \cite{mcnemar1947note}, which is used to compare prediction results on the same test data between two models. 

\noindent\textbf{Reentry Type Prediction}
This task predicts whether the reentry is hateful for hater reentry cases. 
The baseline is calculated by assigning the majority label, i.e., non-hateful, to all samples. 
Table \ref{tab:reentry_type_prediction} shows the evaluation results. 
All models can beat the baseline, however, the Llama 3 zero-shot learning model is very limited in predicting the reentry type (counterspeech F1: 0.53 vs baseline: 0.57). 
Fine-tuned Llama 3 models do not have better results than BERT and BERT-MLT models. 
Similar to the hater reentry prediction, the prediction of reentry type benefits from the conversational context (BERT: hate (0.71) vs counterspeech (0.73) vs pair (0.78), BERT-MTL: hate (0.80) vs counterspeech (0.80) vs pair (82), Llama 3 Finetuned: hate (0.59) vs counterspeech (0.62) vs pair (0.62)), and the differences are significant with McNemar’s tests ($p<0.01$). 


\noindent\textbf{3-way Response Prediction} 
This task predicts one of the three outcomes: hateful reentry, non-hateful reentry, or no reentry. The baseline predicts all samples with ``non-hateful reentry.''
We report the classification results and compare them with the two-stage predictor, which combines the best models, i.e., the BERT-MTL model with pairs, to predict reentry and reentry types. 

Table \ref{tab:three_way_prediction} presents the results. 
Llama 3 zero-shot is almost random at predicting hater reactions. The predictions based on pairs are relatively better than the counterspeech and HS alone, and can slightly beat the majority baseline. 
HS/counterspeech pairs can best predict haters' reactions across all trained models (BERT pair F1 (0.74), BERT-MTL pair F1 (0.77), Llama 3 Zero-shot pair F1 (0.33)), Llama 3 Finetuned pair F1 (0.46)). The two-stage prediction shows an inferior performance (weighted F1: 0.53) to the best 3-way model (weighted F1: 0.77).

\section{Error Analysis}
The best 3-way prediction model still makes errors in some cases. 
We select 200 random samples from the prediction results by the 3-way prediction model and analyze the error cause (annotation details in Appendix \ref{Annotation Details}). Table~\ref{error_analysis} outlines the distribution of common mistakes observed.

Rhetorical question (0.49) causes the most common mistakes. Rhetorical questions are not meant to be answered, but to express strong emotions ~\cite{suzuki2020extracting}, which complex the model understanding. It is also a challenge for HS detection~\cite{schmidt2017survey}. Rhetorical questions account for a high rate of FP and FN across all the classes. 
FN rate in ``no reentry'' (0.63) appears to be the highest, indicating that the model often misinterprets rhetorical questions in the conversation that elicit hateful reentry.

Negation (0.23) is the second most frequent error cause. It often changes the polarity of comments subtly, which is hard for the model to capture. Some HS detection models also struggle with it~\cite{rottger2020hatecheck}. Negation frequently occurs when the model falsely predicts a ``hateful reentry'' (0.29), suggesting that the model may be oversensitive to specific keywords associated with hostility but ignore the negation elements.


Sarcasm and irony (0.17) are also challenging cases similar to other relevant tasks~\cite{grolman2022hateversarial}. Users may use a humorous or satirical tone to express opinions laced with implicit hate~\cite{frenda2022unbearable}. There is a high FP rate (0.27) in ``hateful reentry,'' similar to negation cases, suggesting the model may ignore the reversal meaning in the conversation implied by sarcasm and irony.

Our model also faces challenges with intricate texts (0.10), which use complex syntactic structures to obscure the true intent. In such cases, the model often misclassifies the input as "non-hateful reentry" (0.16). 
Additionally, a few errors occur when the model lacks general knowledge (0.01), such as ``incel'', or ``bigot,'' which may be easier for humans but is challenging for the model.

\section{Conclusion and Discussion}
We introduce a novel way to evaluate different impacts of counterspeech by analyzing hater reactions. 
We have curated the ReEco dataset, consisting of HS/counterspeech and the corresponding hater reactions. 
Our methodology involves linguistic analysis and developing models for predicting hater reactions following counterspeech, which are underexplored in previous research.

We find that certain linguistic features, such as more 2nd person pronouns, and negative words, are tied with hater reentry.
Experiments show incorporating conversation contexts significantly improves the prediction of haters' reactions.
The 3-way prediction method achieves better results than the best two-stage model. 
Our results also show that LLMs have limitations in predicting haters' reactions. Even fine-tuned LLMs are inferior to BERT and BERT-MTL models. 

Our work can be applied practically. First, we provide the ReEco dataset for researchers to analyze the impact of counterspeech on haters' reactions. This dataset includes user-generated counterspeech with various conversation outcomes, which could be used to fine-tune counterspeech generation models, improving their effectiveness in managing hater behavior. Second, our linguistic findings may guide users and counterspeech generation models on how to respond to HS that are less likely to provoke further negative behavior of the hater. Third, we provide insights for social media platforms to assess and intervene in HS. 

\section*{Limitations}
Our study has several limitations. First, we focus on the direct hater reentry following counterspeech in triple-turn conversations, while haters' reactions in the subsequent progression of the conversation were not considered. In future work, we will investigate and categorize more nuanced hater reentry behaviors for a more comprehensive understanding of haters' reactions to counterspeech. Also, the study focuses on the immediate reactions of haters in the conversation, it would also be interesting to examine the impact of counterspeech on haters in the long term. 
Second, models only use HS/counterspeech pairs for predictions. The reentry behavior is also related to users' communication habits, stances, and other conversation contextual factors such as the influence of counter-speakers~\cite{he2021racism}, which are worthy of further investigation. 
Third, in our task of reentry type prediction, we only consider two types of hater reentry. A promising line of future work is to consider the analysis of reentry behaviors with a finer granularity.

\section*{Ethics Statement}

We ensure that our study adheres to ethical guidelines by carefully evaluating associated risks and benefits. 
We collect data from Reddit \footnote{\url{https://pushshift.io/api-parameters/}} under Reddit’s Terms of Service using PushShift API. 
Reddit is a public forum. When users sign up to Reddit, they consent to make their data available to the third party including the academy. Therefore, we can use Reddit data without further seeking user consent following the ethical rules \cite{procter2019study}. 
We have masked users' identifiable information before analysis and modeling.
We will make sure the dataset is exclusively used for non-commercial research purposes\footnote{\url{https://www.reddit.com/wiki/api-terms/}}. 
Examples presented in this study may present toxic languages. We have masked such languages and made notes to readers that they may encounter toxic languages. 
This study aims at effective counter-speech to HS. We acknowledge the potential risks of users being re-identified with anonymized data or misuse of the data by individuals, but the benefits will outweigh such risks. 

\section*{Acknowledgements}
This work was supported by the Institute of Museum and Library Services (IMLS) National Leadership Grants under LG256661-OLS-24 and LG-256666-OLS-24.

\bibliography{custom}
\bibliographystyle{acl_natbib}

\appendix
\label{sec:appendix}
\section{Computing Resources}
\label{Computing Resources}

The neural model takes about half an hour on average
to train in a server with an Intel Xeon Gold 6226R processor, 128 GB memory, and 3 Nvidia RTX 8000 graphic cards.

\section{Hyperparameters of Models}
\label{Hyperparameters of Models}
The hyperparameters for transformer-based models are shown in Table~\ref{table 8}. For Llama-3.1-8B-Instruct zero-shot, we set the temperature as 1.0 and use the default of other hyperparameters. 
The hyperparameters for all tasks with fine-tuned Llama-3.1-8B-Instruct are set to be the same. The num\_train\_epochs is 1, learning rate is $1e^{-5}$, train\_batch\_size is 1, warmup\_ratio is 0.1, lora\_alpha is 32, lora\_dropout is 0.05, and bias is none for the training process, and the top\_k is 50, top\_p is 0.85, do\_sample is true, and repetition\_penalty is 1.0 for the inference process. 

\begin{table}[H]
\resizebox{0.48\textwidth}{!}{
\begin{tabular}{@{}lcccc@{}}
\toprule
Task             & \multicolumn{1}{l}{Epoch} & \multicolumn{1}{l}{Batch size} & \multicolumn{1}{l}{Learning rate} & \multicolumn{1}{l}{Dropout} \\ \midrule
BERT-base        &                           &                                &                                   &                             \\
~~~~~~Hater reentry    & 6                         & 16                             & $1e^{-5}$                                  & 0.1                         \\
~~~~~~Reentry type     & 6                         & 16                             & $1e^{-5}$                                  & 0.1                         \\
~~~~~~3-way & 5                         & 16                             & $1e^{-5}$                                  & 0.1                         \\
BERT-MTL         &                           &                                &                                   &                             \\
~~~~~~Hater reentry    & 7                         & 16                             & $1e^{-5}$                                  & 0.1                         \\
~~~~~~Reentry type     & 7                         & 16                             & $1e^{-5}$                                  & 0.1                         \\
~~~~~~3-way & 5                         & 16                             & $1e^{-5}$                                  & 0.1                         \\
\bottomrule
\end{tabular}}
\caption{Hyperparameters of transformer-based models.}
\label{table 8}
\end{table}

\section{Linguistic Factors}
\label{Linguistic Factors}
We employ SEANCE to conduct linguistic analysis in section 4 on corpus analysis. For most of the linguistic factors, we draw on the explanation provided by \citet{crossley2017sentiment}. To further clarify our findings, we offer a detailed definition of most linguistic factors and references we use to support our results in Table \ref{linguitic factors}.

\begin{table}[ht]
\scalebox{0.58}{
\centering
\begin{tabular}{lp{10cm}}
\hline
\textbf{Factor Type} & \textbf{Definition} \\ 
\hline
\multicolumn{2}{c}{\textbf{Textual Factors}} \\ 

Uncertainty & Words denote feelings of uncertainty \cite{auger2008expression}. \\ 

Abstract & Words indicate a tendency to use abstract vocabulary \cite{borghi2019linguistic}. \\ 

Format & Words refer to formats, standards, or conventions of communication. \\ 

Causation & Words denote the presumption that the occurrence of one phenomenon is necessarily preceded by another \cite{nadathur2020causal}. \\ 
\hline
\multicolumn{2}{c}{\textbf{Emotional Factors}} \\ 

Enlightenment & Words likely reflect a gain in enlightenment through thought, education, etc. \\ 

Polarity & Words show positive or negative sentiment, such as approval, disapproval, or neutrality \cite{abd2021analyzing}. \\ 

Valence & Words describe intrinsic emotions, which can be either positive, negative, or neutral \cite{mohammad2021sentiment}. \\ 

Longing & Words express deep yearning and strong desire. \\ 

Exclamation & Words convey strong sentiments, like surprise, and excitement. \\ 

Aggression & Words indicate anger, frustration, or hostility. \\ 
\hline
\multicolumn{2}{c}{\textbf{Social-related Factors}} \\ 

Respect & Words demonstrate politeness, like please, thank you, and so on. \\ 

Power & Words convey authority, influence, or control in communication \cite{van2016oxford}. \\ 

Worship & Words express adoration or reverence, usually in a religious context \cite{green2017computational}. \\ 

Forgiveness & Words refer to compassion or asking for forgiveness. \\ 
\hline
\end{tabular}}
\caption{Definitions of linguistic factors.}
\label{linguitic factors}
\end{table}

\section{Annotation Details}
\label{Annotation Details}
Our study involves two annotation tasks: HS/ counterspeech identification in Section 3 and error analysis in Section 6. The annotation details in this study are outlined in Table~\ref{Annotation_Details_identification} and Table~\ref{Annotation_Details_error}.  We hire two PhD students with expertise in HS/ counterspeech scope to label the data. Annotators were compensated on average with \$15 per hour. We paid them regardless of whether we accepted their work. Annotators’ IDs are not included in the dataset.
We provide a clear taxonomy: \textbf{HS} refers to expressions in which the author deliberately targets an individual or group with the intent to vilify, humiliate, or incite hatred \cite{yu2022hate}. \textbf{Other comments} refer to statements that do not exhibit such harmful intent, including neutral, supportive, or general discourse. \textbf{Counterspeech} refers to responses to hate speech that are intentionally crafted to contradict and challenge hateful remarks \cite{chung2023understanding}. \textbf{Other replies}, in contrast, encompass responses that do not directly counter hate speech, including neutral, supportive, or unrelated remarks.

In terms of error analysis, we refer to examples from\citet{yu2023hate} to instruct annotators. 
Then 200 instances are randomly selected for validation. Each annotator labels the comments in the same condition but separately. After obtaining the results, the agreement rate and Kappa coefficient are calculated to prove the credibility of human annotation.

\begin{table}[h]
\centering
\scalebox{0.48}{
\renewcommand{\arraystretch}{1.5} 
\begin{tabularx}{\textwidth}{>{\hsize=0.5\hsize}X | >{\hsize=1.5\hsize}X} 
\toprule
\textbf{Task} & \textbf{HS/counterspeech Identification} \\
\midrule
\textbf{Annotator Selection} & Two PhD students with expertise in HS/counterspeech \\
\addlinespace
\textbf{Rule} & \textbf{HS:} Given a comment, annotators determine whether it is HS or not, label "1" as HS, otherwise label "0".\\
& \textbf{Counterspeech:} Given a HS and its reply, annotators determine whether it is counterspeech or not, label "1" as counterspeech, otherwise label "0" \\
\addlinespace
\textbf{Example} & \textbf{HS:} \emph{"Then focus on those other interests? ... incel logic."} \\
& \textbf{Counterspeech:}\emph{ "You're not making any sense. I never said that, that's you saying that."} \\
\addlinespace
\textbf{Credibility} & \textbf{HS:} Agreement rate: 93\%, Kappa coefficient: 0.76; \\
& \textbf{Counterspeech:} Agreement rate: 95\%, Kappa coefficient: 0.79.\\
\bottomrule
\end{tabularx}}
\caption{Annotation details of HS and counterspeech detection.}
\label{Annotation_Details_identification}
\end{table}

\begin{table}[h]
\centering
\scalebox{0.5}{
\renewcommand{\arraystretch}{1.5} 
\begin{tabularx}{\textwidth}{>{\hsize=0.5\hsize}X | >{\hsize=1.5\hsize}X} 
\toprule
\textbf{Task} & \textbf{Error Analysis} \\
\midrule
\textbf{Annotator Selection} & Two PhD students with expertise in HS/counterspeech \\
\addlinespace
\textbf{Rule} & Annotators read the HS and its counterspeech, and label it as "0" for the rhetorical question, "1" for sarcasm or irony; "2" for negation, "3" for general knowledge, and "4" for intricate texts based on error examples. 
\\
\addlinespace
\textbf{Example} & \textbf{Rhetorical Question:} HS: \emph{Just because you can’t be a racist s*** on Twitter doesn’t mean anyone is discriminating against you.} \newline Counterspeech: \emph{Why would you assume anything about race? } \\

& \textbf{Negation:}  HS:\emph{ You are misogynistic. If this is how you speak to a stranger on the internet, I can only imagine how nasty you are in real life.} \newline Counterspeech:\emph{ In reality, it isn't—calling you a worthless b*** isn't misogynistic. [...] You are not "womankind." } \\

& \textbf{Sarcasm or Irony:} HS:\emph{ You are misogynistic. If this is how you speak to a stranger on the internet, I can only imagine how nasty you are in real life.} \newline Counterspeech:\emph{ In reality, it isn't—calling you a worthless b*** isn't misogynistic. [...] You are not "womankind." } \\

& \textbf{Intricate Text:} HS:\emph{ You cannot truly see it from the other perspective either, so please take your half-ass.} \newline Counterspeech: \emph{How can I not see it from your point of view, and you are afraid random women will call you a rapist. That is EXACTLY THE SAME MINDSET AS FEMINISTS WHO hate all men.} \\

& \textbf{General Knowledge:} HS: \emph{Your momma never told you not to stick your d*** in crazy? }\newline Counterspeech: \emph{Incel.}\\

\addlinespace
\textbf{Credibility} & Agreement rate: 91\%; Kappa coefficient: 0.75\\
\bottomrule
\end{tabularx}}
\caption{Annotation details of error analysis.}
\label{Annotation_Details_error}
\end{table}

\section{Subreddit List}
\label{Subreddit List}
We collect Reddit data from the following 42 subreddits and categorize them into five communities: Discussion, Hobby, Identity, Meme, and Media-sharing (Table \ref{Appendix Table 3}). 
\begin{table}[H]
\centering
\scalebox{0.65}{
\begin{tabular}{@{}ll@{}}
\toprule
\textbf{Category} & \textbf{Subreddits}                                                                                                                                                                                         \\ \midrule 
Discussion & \begin{tabular}[c]{@{}l@{}} \textit{r/antiwork, r/changemyview, r/NoFap, r/Seduction,}\\ 
\textit{r/PurplePillDebate, r/ShitPoliticsSays, r/PurplePillDebate,}\\ 
\textit{r/bindingofisaac, r/FemaleDatingStrategy, r/SubredditDrama}\end{tabular} \\
Hobby & \begin{tabular}[c]{@{}l@{}}\textit{r/KotakuInAction, r/DotA2, r/technology, r/modernwarfare,}\\ 
\textit{r/playrust, r/oblivion}\end{tabular} \\
Identity & \begin{tabular}[c]{@{}l@{}}\textit{r/bakchodi, r/Feminism, r/PussyPass, r/MensRights,}\\ 
\textit{r/Sino, r/BlackPeopleTwitter, r/india, r/PussyPassDenied,} \\ 
\textit{r/TwoXChromosomes, r/GenZedong, r/antheism}\end{tabular} \\
Meme & \begin{tabular}[c]{@{}l@{}}r/4Chan, \textit{r/justneckbeardthings, r/HermanCainAward,} \\ 
\textit{r/MetaCanada, r/DankMemes, r/ShitRedditSays}\end{tabular} \\
Media-sharing & \begin{tabular}[c]{@{}l@{}} \textit{r/conspiracy,r/worldnews, r/Drama, r/TumblrInAction,}\\ 
\textit{r/lmGoingToHellForThis, r/TrueReddit.}\end{tabular} \\
\bottomrule
\end{tabular}}
\caption{Subreddit list.}
\label{Appendix Table 3}
\end{table}



\end{document}